%% file: main.tex
\documentclass[journal,twoside,web]{ieeecolor}
\usepackage{jsen}
\usepackage{cite}
\usepackage{amsmath,amssymb,amsfonts}
\usepackage{algorithmic}
\usepackage{graphicx}
\usepackage{textcomp}
\usepackage{wrapfig}

\input{preamble.tex}

\input{commands.tex}
\usepackage{bm}
\newcommand{\citet}[1]{\cite{#1}}
\newcommand{\citep}[1]{\cite{#1}}

\def\BibTeX{{\rm B\kern-.05em{\sc i\kern-.025em b}\kern-.08em
    T\kern-.1667em\lower.7ex\hbox{E}\kern-.125emX}}
\definecolor{abstractbg}{rgb}{0.89804,0.94510,0.83137}
\begin{document}
\title{\titlelong}
\author{Robin Koch$^{1}$, Annabella Mascot$^{2}$, Rayan Younis$^{3,4}$,\\ Martin Wagner$^{3,4}$, Stefanie Speidel$^{4,5}$, Mark Cutkosky$^{2}$, Ingo Sieber$^{6}$, and Roberto Calandra$^{1}$%
\thanks{
This work is supported by the German Research Foundation (DFG) under the Cluster of Excellence CARE: Climate-Neutral And Resource-Efficient Construction (EXC 3115), project number 533767731.
This work is supported by BMFTR in DAAD project 57616814 (\href{https://secai.org/}{SECAI}).
Partially funded by the German Research Foundation (DFG, Deutsche Forschungsgemeinschaft) as part of Germany’s Excellence Strategy – EXC 2050/2 – Project ID 390696704 – Cluster of Excellence “Centre for Tactile Internet with Human-in-the-Loop” (CeTI) of TUD Dresden University of Technology.
This work is supported by the project "Genius Robot" (01IS24083), funded by the Federal Ministry of Education and Research (BMBF).
}\newline
\thanks{$^{1}$ LASR Lab, TUD Dresden University of Technology, Germany\newline
{\tt\small \{rkoch,rcalandra\}@lasr.org}}%
\thanks{$^{2}$Stanford University, Palo Alto, CA}%
\thanks{$^{3}$Department of Visceral, Thoracic, and Vascular Surgery, Faculty of Medicine and University Hospital Carl Gustav Carus, Technische Universität Dresden, Dresden, Germany}%
\thanks{$^{4}$Centre for Tactile Internet with Human-in-the-Loop (CeTI), Technische Universität Dresden, Dresden, Germany}%
\thanks{$^{5}$Department of Translational Surgical Oncology, National Center for Tumor Diseases (NCT), a partnership between DKFZ, University Hospital Carl Gustav Carus, TUD Dresden University of Technology, and Helmholtz-Zentrum Dresden-Rossendorf (HZDR), Dresden, Germany}%
\thanks{$^{6}$ Institut für Automation und angewandte Informatik, Karlsruhe Institute of Technology (KIT), Germany}%
}

\IEEEtitleabstractindextext{%
\fcolorbox{abstractbg}{abstractbg}{%
\begin{minipage}{\textwidth}%
\begin{wrapfigure}[12]{r}{3in}%
\includegraphics[width=3in]{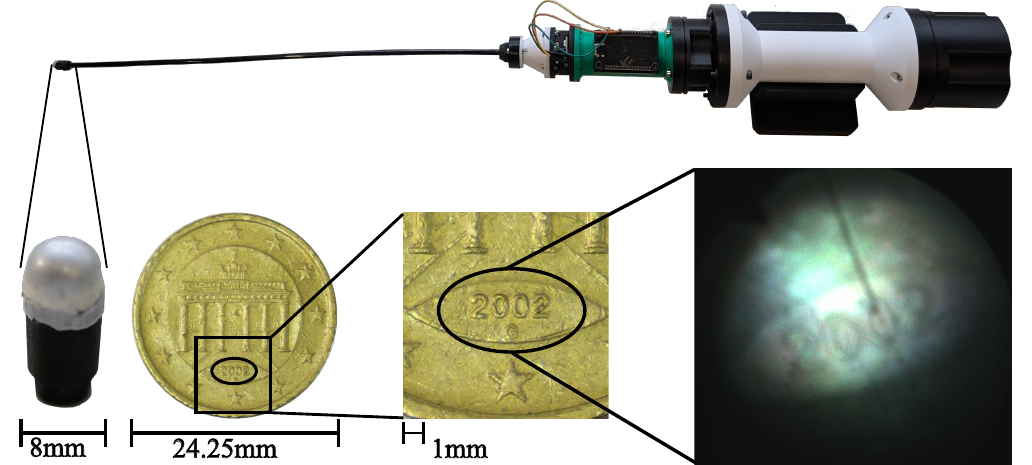}%
\end{wrapfigure}%
\begin{abstract}
	\input{0_abstract.tex}
\end{abstract}

% Uncomment for camera-ready
\end{minipage}}}

\maketitle
%===============================================================================

\section{INTRODUCTION}
	
	\input{1_introduction.tex}
	
%===============================================================================

\section{RELATED WORK}
\label{sec:related}

\input{2_related.tex}

%===============================================================================

\section{Design and Fabrication} 
\label{sec:designandfab}

	\input{3_designAndFab.tex}

%===============================================================================

\section{Sensor Characterization}
\label{sec:sensorCharacterization}

	\input{4_sensorCharacterization.tex}

%===============================================================================

\section{Preliminary Medical Test}
\label{sec:preliminaryMedicalTest}

	\input{5_preliminary_med.tex}

%===============================================================================
\section{Discussion and Future Work}
\label{sec:discussion}

	\input{6_discussion.tex}
%===============================================================================

\section{CONCLUSION}
\label{sec:conclusion}

	\input{7_conclusion.tex}

%===============================================================================

\section*{ACKNOWLEDGMENT}
\input{99_acknowledgments.tex}

%%%%%%%%%%%%%%%%%%%%%%%%%%%%%%%%%%%%%%%%%%%%%%%%%%%%%%%%%%%%%%%%%%%%%%%%%%%%%%%%

\bibliographystyle{IEEEtran}
\bibliography{paper}
\end{document}

%% file: preamble.tex
\usepackage[english]{babel}

\usepackage{graphicx}
\usepackage{animate}
\usepackage{epsfig} 				% for postscript graphics files
\usepackage{epstopdf}

\usepackage{listings}
\usepackage{color}
\usepackage{nameref}
\usepackage{hyperref}
\usepackage{amsmath}	 			% assumes amsmath package installed
\usepackage{amssymb}  				% assumes amsmath package installed

\usepackage{dsfont}			%for the symbol 'Real Numbers'
\usepackage{mathtools}

\usepackage{epigraph}
\usepackage{lscape}
\usepackage[]{nomencl}				% nomenclatures
\usepackage{algorithm}

\usepackage{algorithmic}
\usepackage{multicol}
\usepackage{multirow}
\usepackage{etoolbox}

\usepackage{caption}
\usepackage{subcaption}
\usepackage{wrapfig}

\usepackage{siunitx}

\usepackage{floatflt}

\usepackage{url}

\let\labelindent\relax %allow arxiv to work
\usepackage{enumitem}

\usepackage{tabularx}
\usepackage{placeins}

%% file: commands.tex
\newcommand{\email}[1]{\href{mailto:#1}{\nolinkurl{#1}}}

\newcommand{\fig}[1]{Fig.~\ref{#1}}

\newcommand{\tab}[1]{Table~\ref{#1}}

\newcommand{\namesensor}[0]{MISTac}
\newcommand{\titlelong}[0]{MISTac: A Vision-Based Tactile Sensor for Minimally Invasive Surgery}

\newcommand{\website}[0]{\url{https://github.com/lasr-lab/mistac}}
\newcommand{\forceresolution}[0]{\SI{24.3}{\milli\newton}} 
\newcolumntype{C}{>{\centering\arraybackslash}X}

\newcommand{\thickhline}{%
  \noalign{\global\arrayrulewidth=1.2pt}%
  \hline
  \noalign{\global\arrayrulewidth=0.4pt}%
}
\newcolumntype{V}{!{\vrule width 1.2pt}} 

%% file: 0_abstract.tex
Minimally invasive and robot-assisted  surgery offer many advantages over traditional open surgery, but deprive surgeons of tactile feedback and the ability to palpate tissue with their fingers.
To address this lack of tactile feedback, we introduce the \namesensor{}, a high resolution vision-based tactile sensor specifically designed for palpation in MIS. 
The sensor has a replaceable sensor tip with a diameter of \SI{8}{\milli\meter} which allows it to fit through the trocars used in minimally invasive surgery. 
Its modular 3D-printed case design allows the use of bulky off-the-shelf illumination and imaging hardware that can easily be exchanged and upgraded.
The sensor has an optical resolution of \SI{176.68 }{\micro\meter}, a tactile resolution of \SI{250}{\micro\meter}, and can resolve forces as little as \forceresolution. 
An \textit{in vivo} study with the sensor shows its usability in minimally invasive surgery. 
We trained a machine learning model with the tactile data collected in the trial on a tissue classification task achieving an aggregate accuracy of $\sim$84\% in a leave-one-out cross validation. 
Tactile sensors have the potential to one day aid surgeons during minimally invasive surgery with tasks such as tissue classification or intra-operative tumor localization; \namesensor{} is a small step towards this vision.
We open-source \namesensor{} at \website{}

%% file: 1_introduction.tex
Cancer is one of the leading causes of death worldwide and surgery is frequently the most effective treatment for the removal of localized cancer~\citep{Li2014a}. 
During surgery, complete resection of tumors with adequate resection margins is the main objective to minimize the chance of postoperative tumor recurrence~\cite{Heidkamp2021}. 
Correct resection margins are important for patient survival, as the most frequent cause for the recurrence of malignancies is incomplete removal~\citep{Li2014a}. 
To plan for correct resection margins, accurate localization of the tumor is required during surgery~\citep{Hata2011}. 
Before surgery, tumors are localized using medical imaging technologies such as MRI or CT scans, however, if the tumor is located inside soft and deformable tissue, its position during surgery might shift from the position scanned before the surgery~\citep{Li2014a}. Tumors are often removed "en-bloc", meaning that they are not exposed during resection to prevent tumor cells from spreading in the surgical field. For tumors located completely inside healthy tissue, such as liver tumors in many cases, this has the drawback of inhibiting visual inspection to guide resection.
In open surgery, the surgeon can use their finger to palpate the tissue to locate the hidden tumor during surgery~\cite{Lederman1999}. 
This is not possible during minimally invasive surgery~(MIS) or robot assisted minimally invasive surgery~(RAMIS). Several alternative techniques to locate tumors have been proposed such as intraoperative real-time tissue elastography~\cite{Kobayashi2018} or fluorescence imaging~\citep{Wang2023}. 
However, these alternative techniques come with drawbacks, such as the restricted flexibility of the laparoscopic probe~\citep{Kobayashi2018} or the need for additional markers to be introduced into the patient. 
To aid the intraoperative decision-making process with regards to adequate resection margins, there is still an unmet need for methods that can highlight the margins of the tumor before en-bloc removal and allow for the inspection of the cavity immediately after resection to detect possible residual tumor~\citep{Heidkamp2021}. 

Tactile sensors can be used to digitize touch and offer the opportunity to sense forces, geometries and mechanical properties of the objects with which they are in contact. 
Vision-based tactile sensors~(VBTS)s in particular offer tactile readings with a high spatial resolution that allows to easily detect contours and surface structures.
Their use of digital cameras as transducer elements, furthermore, allows for easy adaption of existing image processing and machine learning techniques and integration with robotic pipelines, thus offering a possible alternative technique that provides tactile readings that can be used for tumor localization and tissue classification during surgery.

In this paper, we present \namesensor{} -- a VBTS with an \SI{8}{\milli\meter} tip diameter which allows it to fit through trocars used in MIS and RAMIS.
The \namesensor{}'s working principle and base architecture design are inspired by the DIGIT Pinki~\citet{Di2025}, upon which we improve by introducing a modular architecture specifically designed for MIS.
\namesensor{} is fully modular at both its distal and proximal end, thus significantly simplifying repair and replacement of parts.
The contributions of this work are as follows:
\begin{itemize}[leftmargin=*,nosep]
    \item We introduce the novel modular design and fabrication of the \namesensor{}, a VBTS that uses bulky off-the-shelf illumination and imaging components while having a sensor tip that is small enough to fit through the trocars used in MIS and RAMIS.
    \item We characterize the \namesensor{}'s optical, tactile and force resolution.
    \item We demonstrate the \namesensor{}'s suitability for MIS in a minimally invasive \textit{in vivo} trial and show its usability in a tissue classification task.
\end{itemize}

%% file: 2_related.tex
\begin{table*}[t]
    \centering
    \renewcommand{\arraystretch}{1.3}
    \renewcommand{\tabularxcolumn}[1]{m{#1}} % vertical centering for X columns
    \setlength{\tabcolsep}{3pt}
    \scriptsize

    \begin{tabularx}{\textwidth}{
        >{\raggedright\arraybackslash}m{1.7cm}
        %|C|C|C|C|C|C|C|C|
        %||C|C|C|C|C|C|C|C
        V C|C|C|C|C|C|C|C
        }
        %\hline
        %\toprule
        \textbf{Sensor/Category} 
        & \textbf{Tip Diameter}
        & \textbf{Optical Resolution}
        & \textbf{Tactile Resolution}
        & \textbf{Force Resolution}
        & \textbf{Flexibility}
        & \textbf{Modularity}
        & \textbf{Housing}
        & \textbf{Dynamic illumination} \\
        %\hline\hline
        %\midrule[1.2pt]
        \thickhline
        DIGIT Pinki~\citet{Di2025}
        & \SI{15}{\milli\meter}
        & \SI{222.72}{\micro\meter}
        & \SI{250}{\micro\meter}
        & \SI{5}{\milli\newton}
        & flexible
        & fully modular, removable tips
        & NA
        & not implemented \\
        \hline
        
        MiniTac~\citep{Li2024a}
        & \SI{8}{\milli\meter}
        & \SI{10}{\micro\meter}\textsuperscript{*}
        & NA
        & \SI{0.6}{\milli\newton}\textsuperscript{**}
        & rigid
        & not modular, tips not replaceable
        & metal shell with carbon handle
        & not possible \\
        \hline

        TacScope~\citep{Prince2025}
        & \SI{7}{\milli\meter}
        & \SI{55.68}{\micro\meter}
        & \SI{250}{\micro\meter}
        & NA
        & rigid
        & modular
        & 3D-printed Shell
        & not possible \\
        \hline

        EndoTac~\citep{Wang2026}
        & \SI{8}{\milli\meter}
        & NA
        & NA
        & \SI{23.97}{\milli\newton}\textsuperscript{**}
        & flexible
        & not modular
        & NA
        & not implemented \\
        \hline

        \namesensor{} (Ours)
        & \SI{8}{\milli\meter}
        & \SI{176.68}{\micro\meter}
        & \SI{250}{\micro\meter}
        & \forceresolution
        & flexible
        & fully modular, removable tips
        & 3D-printed, fully modular
        & possible \\
        %\hline
    \end{tabularx}
    \caption{Comparison of various VBTSs for tissue palpation with the \namesensor{}.
    *Nominal camera resolution, not experimentally verified.
    **Pixel-level force-resolution.}
    \label{tab:ComparisonRelatedMISTac}
\end{table*}

\textbf{Tactile Sensors for Medical Palpation}
Tactile sensors for robotics have been an active field of research for many years, enabling complex manipulation and safe human-robot-interaction through various electrical sensing techniques, including piezoresistive, capacitive, and magnetism-based sensing~\citep{Dahiya2010}. 
Another application of tactile sensors is medical palpation of internal tissue, particularly during MIS and RAMIS where the lack of tactile feedback poses a major challenge to surgeons. 
While many of the tactile sensors developed for use in MIS are aimed at integration into surgery equipment such as grippers and catheters to estimate the applied force of the equipment onto patient tissue to prevent damage by the tools, several sensors were also developed for palpation during MIS for tumor localization specifically~\citep{Othman2022}.  
Modern electronics fabrication enables compact piezoresistive and capacitive transducer arrays on printed circuit boards for integration into forceps~\citep{Ju2024,Arshad2026} and palpation probes~\citep{Naidu2017,Hou2023}.
Tumor localization experiments on \textit{ex vivo} bovine~\cite{Naidu2017} and porcine liver~\citep{Ju2024} as well as porcine kidneys~\citep{Hou2023} with silicone nodules simulating tumors have demonstrated the capability of these sensors to localize tumors.
However, piezoresistive and capacitive transducer arrays usually require individual wiring and multiplexing solutions with increasing transducer array size, thus increasing complexity~\cite{Dong2026}.
Piezoelectric tactile sensors for palpation rely on vibration to detect differences in tissue hardness~\citep{Othman2022}.
Several approaches actively excite the sensor at its resonance frequency and measure its electrical impedance spectrum to detect changes in the resonance frequency when the sensor contacts tissue of different hardness~\citep{Ju2019,Yue2021,Zhang2021}.
Experiments on silicone phantoms and \textit{ex vivo} porcine liver tissue show promising results for tumor localization~\citep{Zhang2021,Ju2019,Yue2021}.
Recent optical-fiber-based palpation sensors often use fiber Bragg gratings~(FBGs), whose reflected Bragg wavelength shifts with axial strain~\citep{Dong2026}. 
FBG-based sensors have been tested on \textit{ex vivo} porcine liver tissue~\citep{Yang2025,Gan2023} and porcine kidney~\citep{Dong2025} with embedded hard objects mimicking tumors, successfully localizing the simulated tumors.
However, FBG sensors can suffer from axis-coupling errors and temperature-induced strain, requiring decoupling and compensation methods~\citep{Dong2026}.
Overall, the described sensor technologies have a low spatial resolution, limited by their transducer density or only measure forces or impedance, requiring rasterized palpation to detect object shapes. 

\textbf{Vision-based Tactile Sensors for Medical Palpation}
VBTSs use a camera to image the deformation of a soft sensor tip, providing high spatial resolution and texture sensing~\citep{Li2025b}.
Sensors such as the DIGIT~\citep{Lambeta2020}, sensors of the GelSight family~\citep{Abad2020} or sensors of the TacTip family~\citep{WardCherrier2018} are commonly mounted on grippers or robotic hands for complex manipulation tasks.
In medicine, VBTSs are an appealing solution for tissue palpation to localize tumors due to their high resolution and sensitivity.
The DIGIT Pinki~\citep{Di2025} is a VBTS with a \SI{15}{\milli\meter} tip designed for digital rectal examinations. 
The sensor design uses a coherent fiber bundle for imaging and incoherent plastic fibers connected to an LED ring for illumination to create a small sensor tip, allowing bulky off-the-shelf optical components to remain at the proximal end.
Notably, the silicone tip is attached via a thread for easy replacement.
The DIGIT Pinki has shown promising results in experiments on silicone phantoms and \textit{ex vivo} tissue~\citep{Di2025}. 
To be employed in RAMIS, however, the sensor tips must fit through the trocars of the robotic system, making miniaturization the main design challenge~\citep{Li2024a}.
The MiniTac~\citep{Li2024a} was developed for tumor detection during RAMIS and designed to fit through the \SI{8}{\milli\meter} trocars of the \textit{Da Vinci Surgical System}. 
Housed in a thin cylindrical metal shell, the sensor features a miniature camera with a silicone tip directly cast onto it.
The tip contains a mechanoresponsive photonic membrane that changes color upon deformation. 
The MiniTac has shown promising results in experiments on silicone phantoms and \textit{ex vivo} tissue~\citep{Li2024a}.
The TacScope~\citep{Prince2025} is a VBTS designed for use in RAMIS. 
The sensor consists of a \SI{7}{\milli\meter} diameter marker-based silicone sensor tip which is attached to a miniaturized camera. 
An additional silicone lens reduces the camera's working distance to make it compatible with the sensor tip.
White LEDs provide the necessary illumination. 
The TacScope achieved promising results for tumor detection in silicone phantoms and was evaluated in a simulated surgical environment~\citep{Prince2025}.
The EndoTac~\citep{Wang2026} is designed for use with a laparoscopic forceps to scan the deformation of vascular structures. 
The sensor consists of a video endoscope and a fiber-based RGB illumination system onto which a silicone tip with a diameter of \SI{8}{\milli\meter} is mounted. 
The tip contains a curved mirror to increase its side-facing sensing surface. 
The EndoTac was tested on commercial and self fabricated artery phantoms, showing promising result in vessel deformation prediction~\citep{Wang2026}.
\tab{tab:ComparisonRelatedMISTac} gives an overview of the sensors and a comparison to the \namesensor{}.
While the MiniTac and the TacScope have tip diameters sized for MIS, their distal ends are rigid and in the case of the MiniTac not easily replaceable.
The EndoTac has a flexible distal end, but requires a laparoscopic forceps to establish contact with tissue and does not have a modular sensor tip.
The DIGIT Pinki offers modularity and a replaceable tip, but is too large for surgical trocars. 

The \namesensor{} aims to combine the advantages of these sensors to form a fully modular VBTS that is flexible and can be built with bulky off-the-shelf components.

%% file: 3_designAndFab.tex
This section describes the design goals of the \namesensor{} and details its design, explaining the function of its individual components and how to fabricate the custom-made parts.
Full schematics of the sensor are available open-source at \website{}.

\subsection{Design Goals}
\label{subsec:DesignGoals}
The design objective of the \namesensor{} is to build a VBTS suitable for MIS leading to the following performance goals:
\begin{enumerate}[leftmargin=*,nosep]
    \item A tip diameter of \SI{8}{ \milli\meter} to fit through the trocars used for minimally-invasive surgery.
    \item An optical resolution of at least \SI{2.52}{\frac{lp}{\milli\meter}} (\SI{232.21}{\micro\meter}) and a tactile resolution of \SI{250}{\micro\meter}.
    \item A fully modular case at proximal end to ensure that individual components can be easily replaced.
    \item A modular sensor tip that is replaceable in between medical procedures or in case of damages.
    \item An illumination system that allows dynamic lighting change via software, as it has been shown~\citep{Redkin2025} that dynamic illumination can significantly improve tactile readings.
\end{enumerate}

\subsection{Overview of the Sensor Architecture}
\label{subsec:overviewSensorArchitecture}
\begin{figure*}[t]
  \centering
  \includegraphics[width=\linewidth]{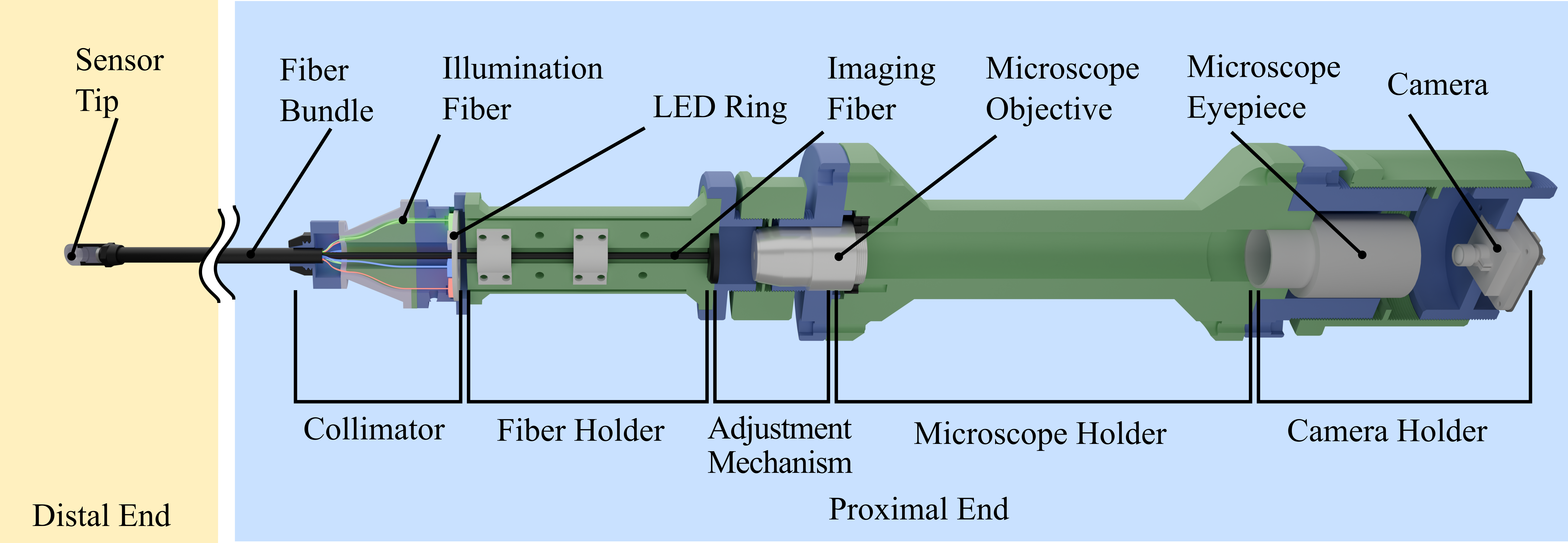}
  \caption{Sensor Overview. The \namesensor{} consists of a gel sensor tip at the distal end and an illumination system and an imaging system located at the proximal end.  The sensor case comprises four modular subassemblies: the collimator, fiber holder, microscope holder, and camera holder. The fiber holder is connected to the microscope holder through an adjustment mechanism. The modular 3D-printed case enables easy repair, customization, and upgrades of the imaging system.}
  \label{fig:fullSensorDesign}
\end{figure*}

The \namesensor{} consists of an imaging system, an illumination system and the sensor tip.
The system, shown in \fig{fig:fullSensorDesign} is divided into two parts: The distal end, pictured on the left and the proximal end, pictured on the right.  
The distal end contains the silicone sensor tip which consists of two layers, an optically clear base layer, and a reflective layer which also acts as the protective layer. 
The sensor tip is attached to an imaging fiber bundle with a \SI{60}{\degree} FOV lens taken from a fiberscope~(\textit{MEDIT}).
An image, formed by the lens, travels along the fiber bundle via total internal reflection towards the proximal end of the sensor. 
A light microscope consisting of a 10x plan-achromat objective~(\textit{KERN OBB-A1238}) and a 10x widefield eyepiece~(\textit{Anti-Fungus OBB-A3200}) magnifies the image.  
The objective has a working distance of \SI{5.65}{\milli\meter} and the assembly together provides a magnification of ${M=100}$. 
A digital camera~(\textit{DFM 37UX568-ML} by \textit{The Imaging Source}) records the images and forms the final part of the imaging system.
An LED ring~(\textit{Adafruit Neopixel Ring}) with 8 controllable RGB LEDs connected to incoherent illumination fibers form the illumination system that provides light at the sensor tip.

\subsection{Design of the Distal End}
\label{subsec:designDistalEnd}
\begin{figure}[t]
    \centering
    \begin{subfigure}[t]{0.3\linewidth}
        \centering        
        \includegraphics[width=1\linewidth, keepaspectratio]{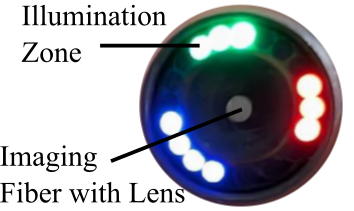}
        \caption{Fiber bundle}
        \label{subfig:illuminationSystem}
    \end{subfigure}
    \begin{subfigure}[t]{0.3\linewidth}
        \centering
        \includegraphics[width=1\linewidth, keepaspectratio]{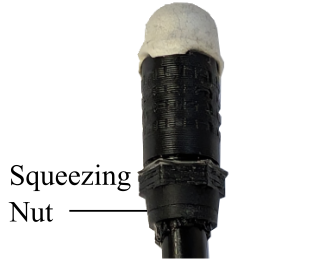}        
        \caption{Tip attached to fiber bundle}
        \label{subfig:tipAttached}
    \end{subfigure}
    \begin{subfigure}[t]{0.3\linewidth}
        \centering
        \includegraphics[width=1\linewidth, keepaspectratio]{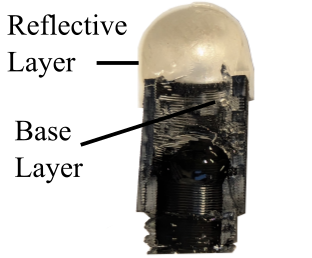}        
        \caption{Tip cross section}
        \label{subfig:tipCrossSection}
    \end{subfigure}
    \caption{Distal End. The built-in illumination fibers of the fiberscope are sorted into 6 illumination zones~(a). This allows to miniaturize the sensor tip which attaches to the fiber bundle and is fastened with a 3D-printed nut~(b). A cross section of the \SI{8}{\milli\meter} sensor tip is shown in~(c).} 
    \label{fig:distalEnd}
\end{figure}

The distal end of the \namesensor{} consists of the sensor tip that is in contact with objects and the distal part of the fiber bundle. 
The fiber bundle consists of a coherent imaging fiber bundle and 19 incoherent illumination fibers, forming the passive part of the imaging and illumination system.
The incoherent illumination fibers require manual sorting to bundle neighboring fibers into groups of 3 or 4 to create a total of 6 illumination zones arranged in a circle around the imaging fiber bundle.
\fig{subfig:illuminationSystem} shows the modified \textit{MEDIT} fiberscope fiber bundle with three of its illumination zones illuminated. 
The fiber bundle has a diameter of \SI{5.6}{\milli\meter} which provides enough design space to add a sensor tip that fits the design constraints.
To fit through the trocars employed in MIS, we designed the silicone sensor tip to have a diameter of \SI{8}{\milli\meter}.
The minimum length of the sensor tip is dictated by the \SI{10}{\milli\meter} working distance of the imaging fibers, which we determined experimentally. 
To allow \SI{2}{\milli\meter} of deformation while remaining within working distance, we set the tip length to \SI{12}{\milli\meter}. 
The sensor tip of the \namesensor{} is elliptical and encased by a 3D-printed cartridge. 
The replaceable cartridge attaches to the fiber bundle with a fastening nut, as shown in \fig{subfig:tipAttached}.
The nut has a conical inner shape that compresses a flexible 3D-printed TPU ring against the fiber bundle, creating a reversible friction fit for quick replacement of the sensor tip. 

\subsection{Fabrication of the Distal End}
\label{subsec:fabricationDistalEnd}
We used 3D-printed molds to cast the base gel of the sensor tip from \textit{Silicone Addition Clear 5A}, an optically clear Pt-curing silicone by \textit{Silicones and more}. 
The silicone has a Shore~A hardness of 5 and is softer than the commonly used \textit{Solaris} silicone by \textit{Smooth On}, which has a Shore~A hardness of 15.
\textit{Silicone Addition Clear 5A} consists of two components A and B, which are mixed in a 1:1 ratio according to the instruction manual. 
The silicone is cast directly into the cartridge which enables the sensor tip to adhere to the cartridge without the need for additional adhesives.
After curing, the base gel is dipped into a cup containing a reflective silicone paint consisting of \textit{CastMagic Silver Bullet} powder by \textit{Smooth-On} mixed with \textit{Ecoflex 00-10} in a 1:10 ratio and is placed onto a flat surface to cure. 
Due to the small tip diameter and its elliptical shape, most of the coating silicone flows off the surface of the base gel producing a very thin and uniform coating layer after curing. 
The excess silicone flows over the cartridge, providing additional adhesion between the cartridge and the gel tip.
\fig{subfig:tipCrossSection} shows a cross section of the sensor tip. 

\subsection{Design and Fabrication of the Proximal End}
\label{subsec:designProximalEnd}
\begin{figure}[t]
    \centering
    \begin{subfigure}[t]{0.3\linewidth}
        \centering        
        \includegraphics[width=1\linewidth, keepaspectratio]{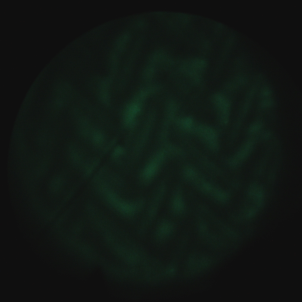}
        \caption{green}
        \label{subfig:dynamicG}
    \end{subfigure}
    \begin{subfigure}[t]{0.3\linewidth}
        \centering
        \includegraphics[width=1\linewidth, keepaspectratio]{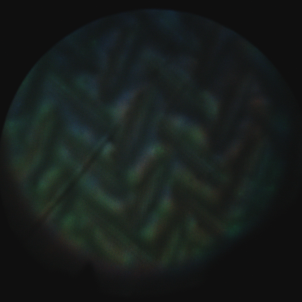}        
        \caption{green, blue}
        \label{subfig:dynamicGB}
    \end{subfigure}
    \begin{subfigure}[t]{0.3\linewidth}
        \centering
        \includegraphics[width=1\linewidth, keepaspectratio]{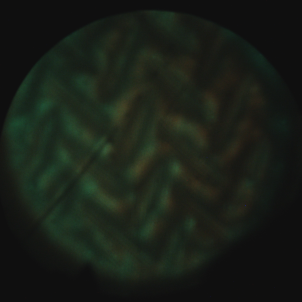}        
        \caption{red, green}
        \label{subfig:dynamicRG}
    \end{subfigure}
       \begin{subfigure}[t]{0.3\linewidth}
        \centering        
        \includegraphics[width=1\linewidth, keepaspectratio]{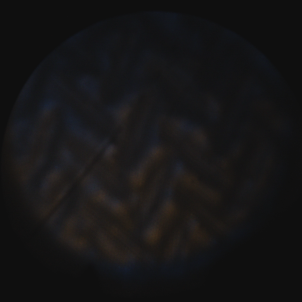}
        \caption{red, blue}
        \label{subfig:dynamicRB}
    \end{subfigure}
    \begin{subfigure}[t]{0.3\linewidth}
        \centering
        \includegraphics[width=1\linewidth, keepaspectratio]{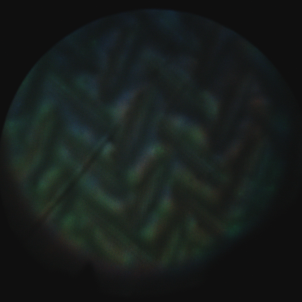}      
        \caption{red, green, blue}
        \label{subfig:dynamicRGB}
    \end{subfigure}
    \begin{subfigure}[t]{0.3\linewidth}
        \centering
        \includegraphics[width=1\linewidth, keepaspectratio]{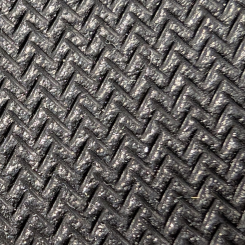}        
        \caption{test object}
        \label{subfig:dynamicTestObject}
    \end{subfigure}
    \caption{Tactile measurements of the \namesensor{} under dynamic illumination. Setting the LEDs to different colors (a)-(e) illuminates the test object from slightly different angles. This highlights different structures of the test object shown in~(f).}
    \label{fig:dynamicIllumination} 
\end{figure}

The proximal end of the \namesensor{} consists of the LED ring, the microscope obejctive and eyepiece as well as the camera which form the active parts of the illumination system and the imaging system respectively.
These components are housed in the sensor case which protects the optical components from stray light. 
The case consists of the following subassemblies: a camera holder, a microscope holder, a fiber holder and a collimator which are shown in \fig{fig:fullSensorDesign}.
The camera holder connects the camera to the rest of the case and includes an adjustment mechanism for translating the camera along the optical axis or removing it to exchange lenses. 
This mechanism uses a turnbuckle with left- and right-handed 3D-printed M33 threads, allowing the camera to move toward or away from the microscope eyepiece without relative rotation.
The microscope holder maintains the required \SI{160}{\milli\meter} spacing between the eyepiece and the objective for sharp imaging. 
The eyepiece slots into one end of the microscope holder and is fixed by a bolt.
The objective is attached at the other end via an RMS thread integrated into a separate 3D-printed disk. 
This disk is used because the required RMS-thread precision could only be achieved reliably in small printed parts. 
The fiber holder is split along the optical axis to simplify insertion, prevent damage to the distal end, and ensure centering of the imaging bundle. 
Two brackets prevent lateral motion, while a flexible TPU centering disk clips around the distal end to provide a tight but reversible friction fit. 
To achieve the working distance of \SI{5.65}{\milli\meter} of the objective and to compensate for assembly tolerances, the fiber holder is connected to the microscope holder using the same adjustment mechanism design that is part of the camera holder.
The collimator connects the LEDs to the illumination fibers and consists of an LED holder that is preceded by an illumination fiber guide.
The fiber guide is split and contains S-shaped channels that guide the illumination fibers toward the holes of the LED holder, reducing the strain on the illumination fibers and simplifying their insertion. 
Each illumination fiber group is interfaced with a different LED of the \textit{Adafruit Neopixel Ring} enabling individual RGB illumination of each group. 
An \textit{ESP32} micro controller connected to a host PC controls the LEDs. 
The LED color and brightness can be adjusted using a Python script which also allows to set dynamic patterns in which individual LEDs are turned on and off, thus creating different illumination patterns. 
\fig{subfig:dynamicG}-\ref{subfig:dynamicRGB} show the different tactile measurements of the test object shown in \fig{subfig:dynamicTestObject} under dynamic illumination.
A flexible TPU fastener, inspired by cable strain reliefs, attaches to the illumination fiber guide and fixes the combined fiber bundle. 
The fastener consists of a disk with a partially threaded hollow cylinder.
Six slits allow the end without thread to compress when an M11 nut with a conical inner hole is tightened, creating a friction fit around the bundle.
All case components are 3D-printed and connected using nuts, bolts or 3D-printed threads. 
This modular design simplifies repair and component replacement, such as camera upgrades, without requiring a complete case redesign.

%% file: 4_sensorCharacterization.tex
This section describes the experiments performed to evaluate the \namesensor{}'s performance. 
\subsection{Sensor Resolution}
\label{subsec:SensorResolution}
\begin{figure}[t]
  \centering
    \begin{subfigure}[t]{0.49\linewidth} % Left image
        \centering        
        \includegraphics[width=1\linewidth, keepaspectratio]{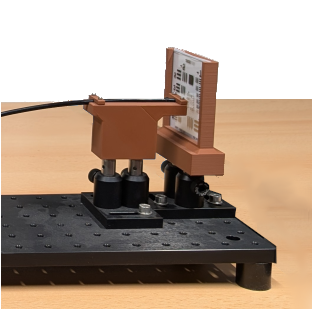}
        \caption{Optical Resolution Testbench}
        \label{subfig:testSetup}
    \end{subfigure}
    \begin{subfigure}[t]{0.49\linewidth} % right image
        \centering        
        \includegraphics[width=1\linewidth, keepaspectratio]{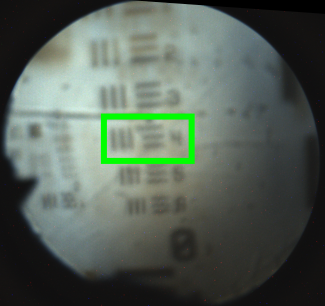}
        \caption{Optical Resolution}
        \label{subfig:opticalResolution}
    \end{subfigure}
        \begin{subfigure}[t]{0.49\linewidth} % Left image
        \centering        
        \includegraphics[width=1\linewidth, keepaspectratio]{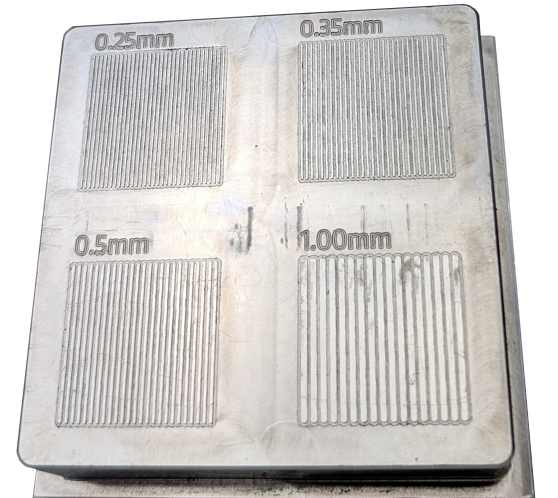}
        \caption{Aluminum Testblock}
        \label{subfig:aluBlock}
    \end{subfigure}
    \begin{subfigure}[t]{0.49\linewidth} % right image
        \centering        
        \includegraphics[width=1\linewidth, keepaspectratio]{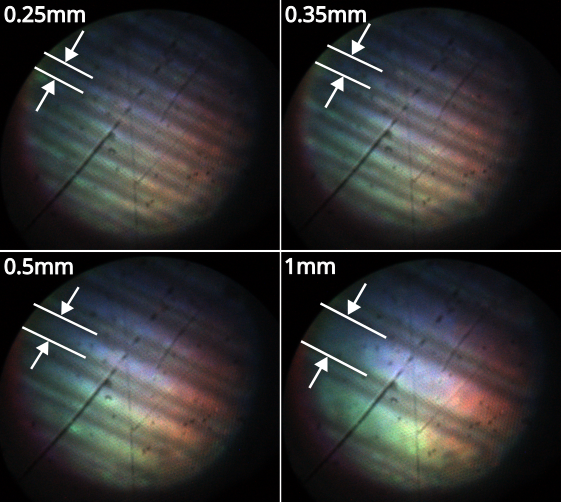}
        \caption{Tactile Resolution}
        \label{subfig:tactileOutput}
        \end{subfigure}
  \caption{The distal end of the \namesensor{} is mounted on a test bench made from an optical breadboard and 3D-printed parts, where it images a USAF~1951 resolution target~(a). The resulting image resolves element 4 of group 1, corresponding to an optical resolution of \SI{176.68}{\micro\meter}. A scratch caused by fiber damage during assembly is visible and extends from the edge toward the green rectangle~(b). Tactile resolution is evaluated using a milled aluminum block with line pairs at four different spacings~(c). The tactile output shows that the \namesensor{} has a tactile resolution of \SI{250}{\micro\meter}~(d).}
  \label{fig:testResolution}
%  \vspace{-10pt}
\end{figure}

We designed a test bench consisting of 3D-printed parts and an optical breadboard to perform measurements of the optical resolution using a USAF~1951 resolution target from \textit{Thorlabs}. 
\fig{subfig:testSetup} shows the test bench. 
A 3D-printed part holds the \namesensor{}'s distal end with its sensor tip removed and positions it \SI{12}{\milli\meter} away from the resolution target so that the distance corresponds to the length of the sensor tip.
The \namesensor{}'s camera takes images of the resolution target which are then evaluated to determine which element of the target is still visible.
\fig{subfig:opticalResolution} shows the resolution target imaged by the \textit{DFM 37UX568-ML} with a \SI{70}{\degree} FOV lens attached. 
The green rectangle marks the bars of element 4 of group 1 that are properly resolved. 
This yields an optical resolution of \SI{176.68}{\micro\meter} which corresponds to a resolution of 2.83~$lp/mm$.
Following the method reported in~\citet{Di2025}, we determined the \namesensor{}'s tactile resolution using a machined aluminum test block.
The block, shown in \fig{subfig:aluBlock}, has four different sections containing line pairs spaced \SI{1}{\milli\meter}, \SI{0.5}{\milli\meter},\SI{0.35}{\milli\meter}, \SI{0.25}{\milli\meter} apart. 
The \namesensor{}'s sensor tip is pressed against the different sections of the test block, tactile readings are recorded and subsequently evaluated to determine which line pairs are still resolved. 
The tactile resolution of the \namesensor{} is at least \SI{250}{\micro\meter}, as shown in \fig{subfig:tactileOutput} where the line pairs are clearly visible.
\subsection{Force Resolution}
\label{subsec:NormalAndShearForceResolution}
\begin{figure}[t]
  \centering
    \begin{subfigure}[t]{0.49\linewidth} % Left image
        \centering        
        \includegraphics[width=1\linewidth, keepaspectratio]{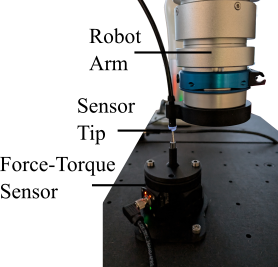}
        \caption{Data collection setup}
        \label{subfig:testSetupForce}
    \end{subfigure}
    \begin{subfigure}[t]{0.49\linewidth} % right image
        \centering        
        \includegraphics[width=1\linewidth, keepaspectratio]{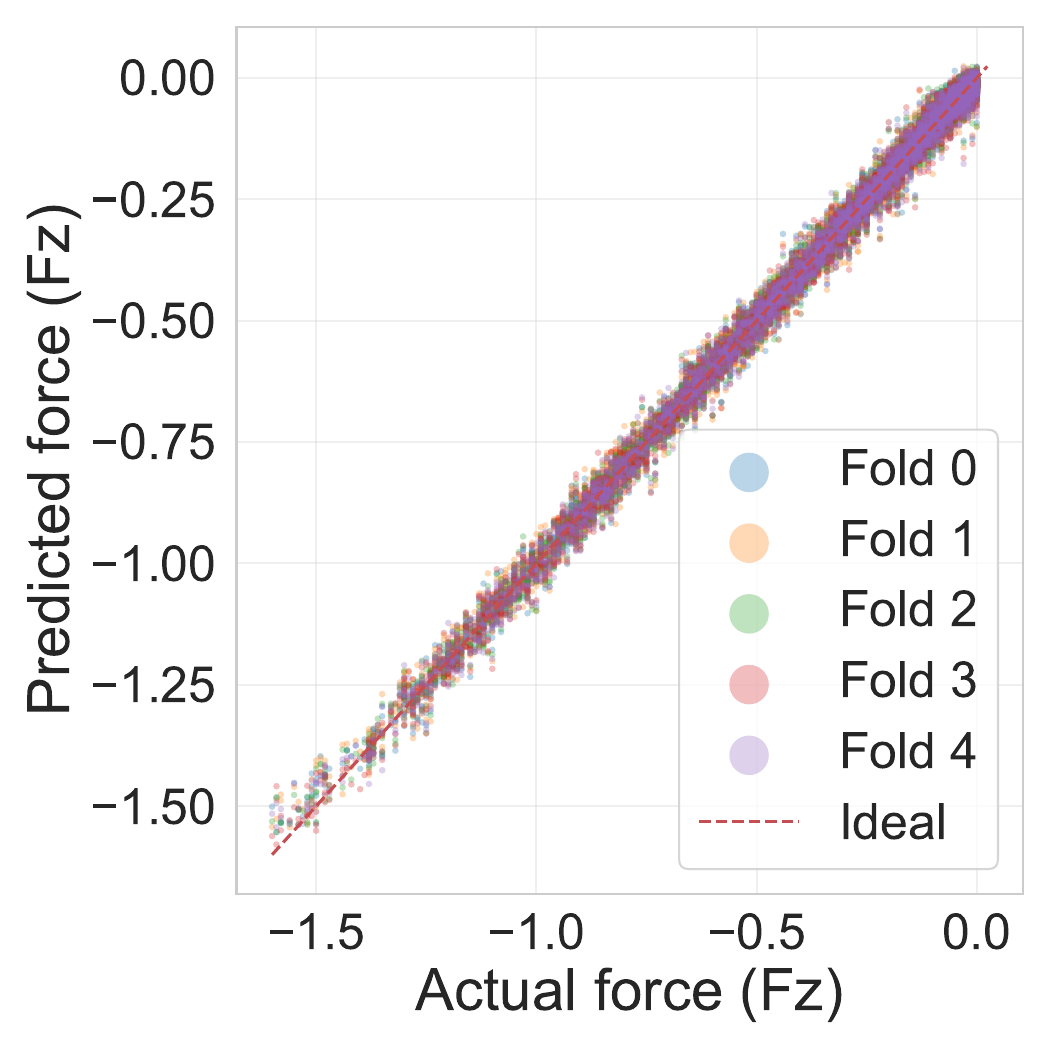}
        \caption{Predicted vs. actual Fz [N]}
        \label{subfig:normalForcePrediction}
    \end{subfigure}
  \caption{Our data-collection setup consists of an x-arm 7 robot arm with the \namesensor{}'s distal end attached. While recording tactile readings, the robot presses the sensor tip against a \SI{1.2}{\milli\meter} metal probe mounted on a BotaSys SensOne fts which provides the ground truth force values~(a). The predicted vs. actual normal force [N] is shown in~(b). Our modified ResNet18 model achieves an RMSE of \SI{24.3}{\milli\newton} ± \SI{0.9}{\milli\newton}.}
  \label{fig:forceResolution}
\end{figure}

We used methods similar to those reported by~\citet{Di2025} to obtain a normal force resolution under a controlled loading condition. 
While this loading condition is not representative of the loading that will occur in practice when the device is used for laparoscopic palpation, it provides a force sensitivity measurement that allows comparison with other sensors.
Our data collection setup, shown in \fig{subfig:testSetupForce}, consists of a robotic arm, a \textit{Botasys SensONE Gen 0 BFT-SENS-ECAT-M8} force torque sensor~(FTS) and the \namesensor{}. 
The \textit{ufactory x-arm 7} robot arm presses the \namesensor{}'s sensor tip onto a metal probe that has a diameter of \SI{1}{\milli\meter}. 
The probe is attached to the fts that provides ground truth force data and has a noise level of \SI{20}{\milli\newton}.
To create the normal force data set, the sensor tip is probed at 5 different points in a \SI{2}{\milli\meter} by \SI{2}{\milli\meter} area, while recording tactile readings.  
The robot arm presses the sensor tip down onto the metal probe at random depths while applying forces within a \SI{1.6}{\newton} limit. 
The force data and the tactile data are sampled at \SI{50}{\hertz} and \SI{30}{\hertz}, respectively. 
We pair force data and tactile readings by matching their timestamps, thus creating pairs with a time difference of less than \SI{5}{\milli\second}, resulting in a total of 49,201 pairs.
For force regression from tactile readings, we employed a ResNet-18~\citep{He2016} architecture pretrained on ImageNet~\citep{Deng2009}, with a 350×350×3 input layer, and modified the final linear layer to produce a single scalar output representing the estimated force value. 
We preprocessed images using a center crop followed by downscaling to match the model's input dimensions; no additional preprocessing was required.
We fine-tuned the model for up to 90 epochs with a batch size of 32, applying early stopping with a patience of 10 epochs on the validation loss. 
We initialized the learning rate at $1.2\times10^{-5}$ and adjusted it using a cosine annealing scheduler.
We used mean square error between predicted and ground-truth force values as the loss function.
To evaluate generalization, we conducted five-fold cross-validation on the normal force dataset, achieving a root mean squared error~(RMSE) of \SI{24.3}{\milli\newton} ± \SI{0.9}{\milli\newton}, which is consistent with the noise floor of the reference sensor used for labeling. 
\fig{subfig:normalForcePrediction} shows the plot of predicted vs. actually measured normal forces for the five folds.
The prediction for all folds clearly follow the ideal line, indicating consistent force estimation. 

%% file: 5_preliminary_med.tex
To evaluate the \namesensor{}'s employability in a medical context, we tested it in a minimally invasive\textit{ in vivo} study on n=2 sedated pigs. All procedures were approved by the local state authority with the  approval number TVV43/2023 (Saxony, Germany), and conducted in accordance with institutional ethical standards for animal experimentation and the registered protocol.   
The study was performed with an older version of the \namesensor{} that used an \textit{Arducam B0268} as its digital camera.

\subsection{Data Collection}
\label{subsec:DataCollection}
\begin{figure}[t]
  \centering
  \includegraphics[width=1\linewidth]{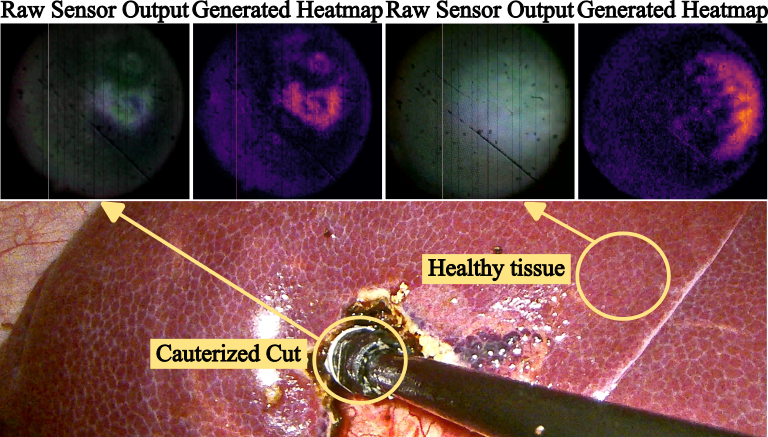}
  \caption{The \namesensor{} used in an minimally invasive \textit{in vivo} trial. The bottom image shows endoscope footage of the sensor tip touching a cauterized liver cut. 
  The raw sensor output in the top left clearly shows the cut surface. 
  Healthy liver tissue, shown in the top right, produces only subtle features in the raw output.
  Subtracting a neutral image and normalizing the output highlights the surface features in the corresponding heatmaps.
  }
  \label{fig:medicalTrial}
\end{figure}

To collect the tactile data during surgery, a surgeon inserted the sensor through one trocar, while an assistant controlled an endoscope through another trocar.
Oriented by the endoscope footage, the surgeon palpated different organ tissue with the \namesensor{}, and we recorded an audio track in which the surgeon identified the tissue contacted for labeling. 
Using this method, we collected \SI{40}{\minute} of tactile footage at 30 fps. 
Due to its modular design, the sensor tip used could easily be discarded after the procedure and replaced by a fresh one. 
\fig{fig:medicalTrial} shows endoscope footage of the \namesensor{} touching a cauterized cut of the liver with the frame-matched raw sensor output shown in the top left.
While the liver cut is clearly visible in the raw sensor output, healthy liver tissue shown in the top right produces only very subtle features when the sensor tip slides over it. 
To make these features more apparent, we subtract a neutral tactile reading captured without tissue contact from the contact images and normalize the result across all three color channels, allowing the features to be visualized as a heatmap.
\subsection{Tissue Classification}
\label{subsec:TissueClassification}
\begin{figure}[t]
    \centering        
    \includegraphics[width=0.98\linewidth, keepaspectratio]{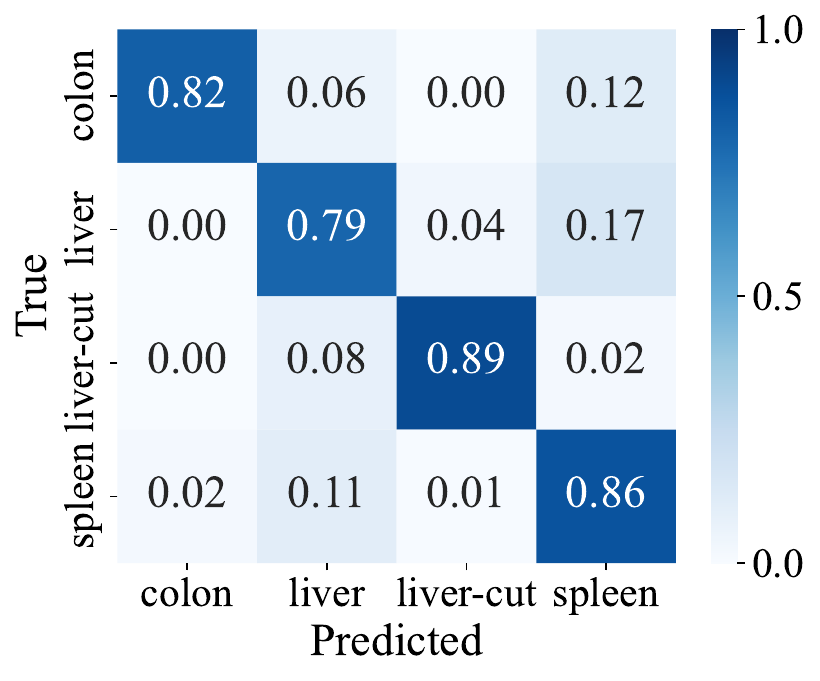}
\caption{Row-normalized confusion matrix after leave-one-out cross validation. Our model classifies tactile readings of colon, healthy liver tissue, a cauterized liver cut and spleen. The results indicate that the \namesensor{}'s tactile readings can be used to perform tissue classification tasks.}
  \label{fig:confusionMatrix}
\end{figure}

To evaluate whether the tactile data can be used for tissue classification, we trained a machine learning model to classify four different tissue types: liver, cauterized liver cut, spleen and colon. 
The tactile footage of one of the \textit{in vivo} trials was labeled using the recorded audio annotations and split into clips for each tissue class yielding a total of 72 clips. 
The dataset is imbalanced, with 11 colon clips with a total of 3,613 frames, 17 spleen clips with a total of  6,080 frames, 21 liver clips with a total of 9,107 frames and 23 cauterized liver-cut clips with a total of  8,210.
The clip lengths vary substantially, especially for spleen and liver-cut.
To exploit temporal information from the tactile data, we trained a pretrained ResNet18 feature extractor followed by a small causal Transformer encoder~\citep{Vaswani2017} using the decoder-only architecture of~\citet{Radford2018}. 
We split the dataset in a train, validation and a test set at the clip level, to avoid overlap between splits.
Training sequences were generated with an eight-frame sliding window, while validation and test sequences used a 16-frame window to prevent overlap. 
Our preprocessing pipeline crops the images and rescales them to the 224x224 input size of the ResNet and performs crop and shift, as well as horizontal and vertical flip to all images of the frame sequence to augment the data. 
We trained the model using the Adam optimizer with a one cycle scheduler, early stopping with 20 epochs patience, and a weighted sampler to balance the classes.
To account for the dataset imbalance, we used leave-one-out cross validation to test our model.
The model achieved an aggregate accuracy of 0.8434, Macro F1-score of 0.8473, Macro Precision of 0.8586, and Macro Recall of 0.8427.
\fig{fig:confusionMatrix} shows the aggregate row-normalized confusion matrix.
The model confuses liver and spleen samples most often, with spleen having the weakest F1 score of 0.7796 and liver the lowest recall of 0.7931.
The results show that the \namesensor{} yields tactile readings of soft tissue that can be used for tissue discrimination tasks. 

%% file: 6_discussion.tex
\subsection{Comparison to other Sensors}
\tab{tab:ComparisonRelatedMISTac} compares the \namesensor{} presented in this work with the DIGIT Pinki~\citep{Di2025}, the MiniTac~\citep{Li2024a}, the TacScope~\citep{Prince2025} and the EndoTac~\citep{Wang2026}. 
The \namesensor{} has a distal end with a diameter of \SI{8}{\milli\meter} and is, like the MiniTac, the TacScope, and the EndoTac, small enough to fit through the \SI{8}{\milli\meter} trocars used in MIS. 
The \namesensor{} has the same tactile resolution as the DIGIT Pinki and the TacScope and an optical resolution of \SI{176.68}{\micro\meter} which is slightly better than the DIGIT Pinki's optical resolution. 
Due to the fiber optics employed in the system, the optical resolution is, however, still considerably worse than the reported optical resolution of \SI{10}{\micro\meter} and  \SI{55.68}{\micro\meter} of the MiniTac and the TacScope respectively. 
With a force resolution of~\forceresolution{} the \namesensor{} is roughly on par with the reported pixel-level force resolution of the EndoTac (\SI{23.97}{\milli\newton}), but has a worse performance than the EndoTac's learning-based MAE (\SI{8}{\milli\newton}),  DIGIT Pinki and the MiniTac which have a reported force resolution of \SI{5}{\milli\newton} and \SI{0.6}{\milli\newton} respectively.
Like DIGIT Pinki and EndoTac, \namesensor{} has a flexible distal end, whereas the MiniTac and TacScope are completely rigid. 
The \namesensor{}, the DIGIT Pinki and the TacScope have a removable sensor tip, which provides a modularity that is lacking in the MiniTac and EndoTac.  
Full modularity is also provided in the housing of the \namesensor{} at the proximal end allowing replacement of all sensor components thus facilitating the repairing and upgrading individual sensor parts. 
The housing of the MiniTac and the Tacscope are not modular and the DIGIT Pinki does not have a case.
The proximal end of the EndoTac is not described by the authors.
The software interface designed for the \namesensor{} provides the user with the option to select dynamic LED settings which allows dynamic RGB illumination of the sensor tip. 
While the DIGIT Pinki's and the EndoTac's hardware in principle allows for dynamic RGB lighting it is not implemented. 
Due to their different working principle, the MiniTac and TacScope do not use RGB illumination and therefore dynamic RGB illumination is not possible. 

\subsection{Limitations and Future Work}
The \namesensor{}'s design has several limitations.
Firstly, the imaging fiber bundle limits the optical resolution: the current fiber bundle contains 7400 cores, whereas high-end medical fiber bundles often contain more than twice as many. 
These were not used due to their high cost, as this setup was intended as a first proof of principle.
In addition, the current fiber bundle was damaged during assembly, causing visual artifacts. 
A future design of the \namesensor{} should therefore use a high-end fiber bundle to increase the optical resolution which should be straightforward to integrate due to the modular design.
The preliminary medical trial also showed that an actuated distal end would help reach regions that are difficult to access because of trocar placement.
Since laparoscopic endoscopes of similar size already use cable-driven bending sections, this could be implemented in a future \namesensor{} design.
The modular housing also allows a potential adaption of the sensor to the \textit{Da Vinci Surgical System (Intuitive Inc., Sunnyvale, USA)} without requiring a complete redesign.
In the preliminary medical trial, we sampled tissue that was already clearly distinguishable using an endoscope.
Thus, while the tactile data and tissue-classification task provide a useful proof of concept, future studies should target tissues where tactile readings of the \namesensor{} offer greater surgical benefit.
Promising directions include lump or blood-vessel detection. 
Finally, future work could leverage the \namesensor{}'s high spatial resolution to create local 3D maps of organs or organ regions from tactile measurements.

%% file: 7_conclusion.tex
In this work we present \namesensor{}, a fully modular miniaturized VBTS designed for MIS and RAMIS.
\namesensor{} uses bulky off-the-shelf components while keeping a small tip diameter which allows it to be inserted through trocars.
Experimental evaluation shows that \namesensor{} has an optical resolution of~\SI{176.68}{\micro\meter}, a tactile resolution of~\SI{250}{\micro\meter} and a force resolution of~\SI{24.3}{\milli\newton}.
In an \textit{in vivo} trial we showed that the sensor can be used in MIS.
We trained a machine learning model using the tactile data collected in the \textit{in vivo} trial to solve a proof-of-concept tissue classification task, highlighting \namesensor{}’s potential for future medical applications 

%% file: 99_acknowledgments.tex
We thank the team of the Experimental Operating Room at the NCT Dresden for their support. We thank the Zentrum für Informationsdienste und Hochleistungsrechnen (ZIH) at TU Dresden for providing computing resources. 
\textbf{Author's contributions}.
R.C. and R.K. conceptualized the sensor. R.C. and I.S. supervised the project. R.K. designed and manufactured the sensor, and performed the sensor characterization. R.K., A.M., R.Y., and M. W. collected the medical data. R.K. wrote the initial draft. A.M., R.Y., M.W., S.S., M.C., I.S. and R.C. reviewed and edited the paper.    
\FloatBarrier